%% file: main.tex
\documentclass[10pt,twocolumn,letterpaper]{article}
\usepackage[accsupp]{axessibility}  

\usepackage[pagenumbers]{cvpr} 

\input{preamble}

\definecolor{cvprblue}{rgb}{0.21,0.49,0.74}
\usepackage[pagebackref,breaklinks,colorlinks]{hyperref}
\definecolor{citecolor}{rgb}{0.8, 0.4, 0.9}
\definecolor{linkcolor}{HTML}{ED1C24}
\hypersetup{
  colorlinks=true,
  citecolor=cvprblue,
  linkcolor=linkcolor,   
}

\def\paperID{16585} 
\def\confName{CVPR}
\def\confYear{2025}

\newcommand{\ours}{DiffusionSfM}

\title{DiffusionSfM: Predicting Structure and Motion \\ via Ray Origin and Endpoint Diffusion}

\author{Qitao Zhao, Amy Lin, Jeff Tan, Jason Y. Zhang, Deva Ramanan, Shubham Tulsiani \\
Carnegie Mellon University\\
\textit{Project page:} \href{https://qitaozhao.github.io/DiffusionSfM}{qitaozhao.github.io/DiffusionSfM}\\
}

\begin{document}
\maketitle
\input{sec/0_abstract}    
\input{sec/1_intro}
\input{sec/2_related_work}
\input{sec/3_method}

\input{sec/4_experiments}

\input{sec/5_discussion}

\newpage
\noindent \textbf{Acknowledgements:} We thank the members of the Physical Perception Lab at CMU for their valuable discussions, and extend special thanks to Yanbo Xu for his insights on diffusion models.

This work used Bridges-2 \cite{brown2021bridges} at Pittsburgh Supercomputing Center through allocation CIS240166 from the Advanced Cyberinfrastructure Coordination Ecosystem: Services \& Support (ACCESS) program, which is supported by National Science Foundation grants \#2138259, \#2138286, \#2138307, \#2137603, and \#2138296. This work was supported by Intelligence Advanced Research Projects Activity (IARPA) via Department of Interior/Interior Business Center (DOI/IBC) contract number 140D0423C0074. The U.S. Government is authorized to reproduce and distribute reprints for Governmental purposes notwithstanding any copyright annotation thereon. Disclaimer: The views and conclusions contained herein are those of the authors and should not be interpreted as necessarily representing the official policies or endorsements, either expressed or implied, of IARPA, DOI/IBC, or the U.S. Government.

{
    \small
    \bibliographystyle{ieeenat_fullname}
    \bibliography{main}
}

\input{sec/X_suppl}

\end{document}

%% file: preamble.tex
\usepackage{amsmath,amsfonts,bm}
\usepackage{multirow}
\usepackage{multicol}
\usepackage{cuted}
\usepackage{soul}

\def\eqref#1{equation~\ref{#1}}

\def\1{\bm{1}}

\DeclareMathAlphabet{\mathsfit}{\encodingdefault}{\sfdefault}{m}{sl}
\SetMathAlphabet{\mathsfit}{bold}{\encodingdefault}{\sfdefault}{bx}{n}



%% file: sec/0_abstract.tex
\begin{strip}
\centering
\vspace{-40pt}
\includegraphics[width=1.0\textwidth]{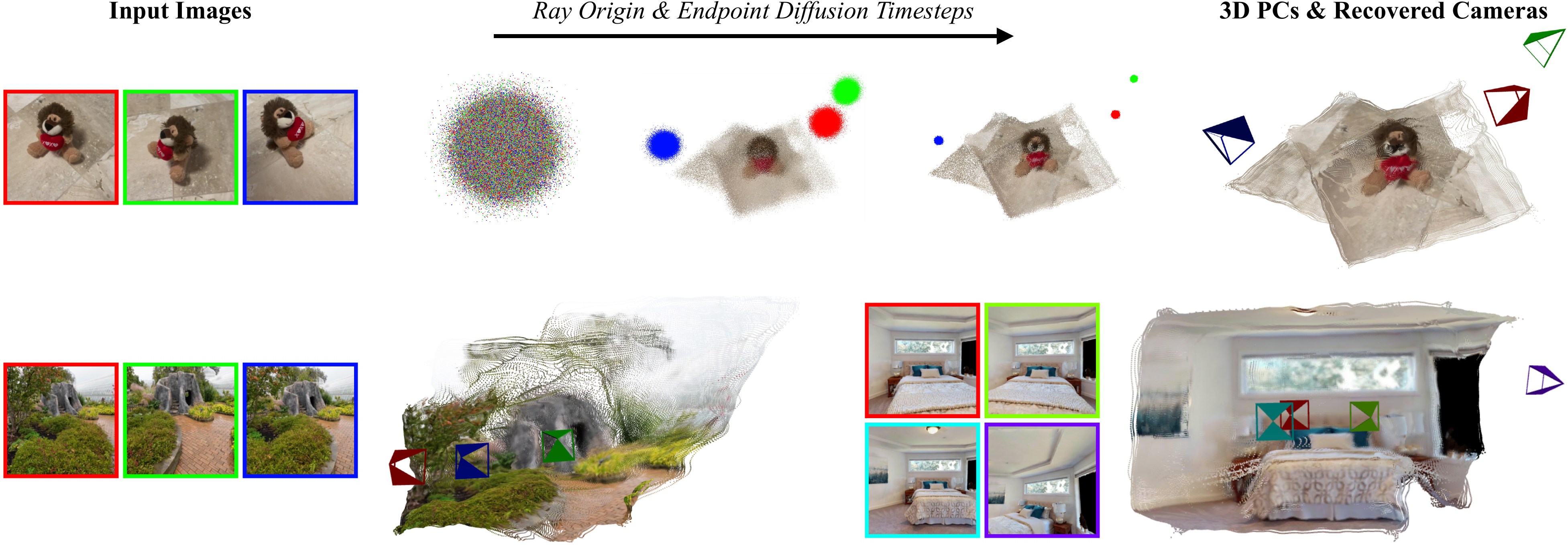}
\captionsetup{hypcap=false}\captionof{figure}{\textbf{DiffusionSfM.} \textbf{Top:} Given a set of multi-view images (left), DiffusionSfM represents scene geometry and cameras (right) as pixel-wise ray origins and endpoints in a global frame. It learns a denoising diffusion model to infer these elements directly from multi-view inputs. Unlike traditional Structure-from-Motion (SfM) pipelines, which separate pairwise reasoning and global optimization into two stages, our approach unifies both into a single end-to-end multi-view reasoning framework.  
\textbf{Bottom:} Example results of inferred scene geometry and cameras for two distinct settings: a real-world outdoor scene (left) and a synthetic indoor scene (right).}
\label{fig:teaser}
\end{strip}

\begin{abstract}
Current Structure-from-Motion (SfM) methods typically follow a two-stage pipeline, combining learned or geometric pairwise reasoning with a subsequent global optimization step. In contrast, we propose a data-driven multi-view reasoning approach that directly infers 3D scene geometry and camera poses from multi-view images. Our framework, \ours, parameterizes scene geometry and cameras as pixel-wise ray origins and endpoints in a global frame and employs a transformer-based denoising diffusion model to predict them from multi-view inputs. To address practical challenges in training diffusion models with missing data and unbounded scene coordinates, we introduce specialized mechanisms that ensure robust learning. We empirically validate \ours\ on both synthetic and real datasets, demonstrating that it outperforms classical and learning-based approaches while naturally modeling uncertainty.
\end{abstract}

%% file: sec/1_intro.tex
\section{Introduction}
\label{sec:intro}

The task of recovering structure (geometry) and motion (cameras) from multi-view images has long been a focus of the computer vision community, with typical pipelines~\cite{schonberger2016structure} performing pairwise correspondence estimation followed by global optimization. While classical methods relied on hand-designed features, matching, and optimization, there has been a recent shift towards incorporating learning-based alternatives~\cite{detone2018superpoint,sarlin2020superglue,lindenberger2021pixsfm,cai2023doppelgangers}. More recently, the widely influential DUSt3R~\cite{wang2023DUSt3R} advocates for predicting pairwise 3D pointmaps (instead of only correspondences), demonstrating that this can yield accurate dense geometry and cameras. In order to reconstruct more than two views, DUSt3R (and its variants \cite{leroy2024grounding}) still require a global optimization reminiscent of classic bundle adjustment. While these methods, both classical and learning-based, have led to impressive improvements in SfM, the overall approach is largely unchanged -- learned or geometric pairwise reasoning followed by global optimization. In this work, we seek to develop an alternative approach that directly predicts both structure and motion, unifying pairwise reasoning and global optimization into a single multi-view framework.

We are of course not the first to attempt to find unified alternatives to the two-stage SfM pipeline. In the sparse-view setting where conventional correspondence-based methods struggle, several works employ multi-view architectures to jointly reason across input images. SparsePose~\cite{sinha2023sparsepose}, RelPose++~\cite{lin2023relpose++}, and PoseDiffusion~\cite{wang2023posediffusion} all leverage multi-view transformers to estimate camera pose for input images, albeit using differing mechanisms such as regression, energy-based modeling, and denoising diffusion. More recently, RayDiffusion~\cite{zhang2024cameras} argues for a local raymap parameterization of cameras instead of a global extrinsic matrix and shows that existing patch-based transformers can be easily adapted for this task, yielding significantly more accurate pose predictions. Importantly, such methods predict only camera motion and fail to predict scene structure.

In this work, we present \ours, an end-to-end multi-view model that directly infers dense 3D geometry and cameras from multiple input images. Instead of inferring (depth-agnostic) rays per image patch (as in RayDiffusion~\cite{zhang2024cameras}) or 3D points per pixel (as in DUSt3R~\cite{wang2023DUSt3R}), \ours\ effectively combines both to predict ray \emph{origins and endpoints} per pixel, directly reporting both scene geometry (endpoints) and generalized cameras (rays). These can readily be converted back to traditional cameras~\cite{zhang2024cameras}. Compared to RayDiffusion, our model directly predicts structure as well as motion at a finer scale (pixel-wise v.s. patch-wise). Compared to DUSt3R, our model directly predicts motion as well as structure but, even more importantly, does so for $N$ views, eliminating the need for memory-intensive global alignment. To model uncertainty, we train a denoising diffusion model but find two key challenges that need to be addressed. First, diffusion models require (noisy) ground truth as input for training, but existing real datasets do not have known endpoints for all pixels due to missing depth in multi-view stereo. Second, the 3D coordinates of endpoints can be potentially unbounded, whereas diffusion models require normalized data. We develop mechanisms to overcome these challenges, leveraging additional ``GT mask conditioning'' as input to inform the model of missing input data and parameterizing 3D points in projective space instead of Euclidean space. We find these strategies allow us to learn accurate predictions for structure and motion.

We train and evaluate \ours\ on real-world and synthetic datasets \cite{reizenstein2021common, savva2019habitat, zhou2018stereo} and find that it can infer accurate geometry and cameras for both object-centric and scene-level images (see Fig.~\ref{fig:teaser}). In particular, we find that \ours\ yields more accurate camera estimates compared to prior work across these settings while also modeling the underlying uncertainty via the diffusion process. In summary, we show that \ours\ can serve as a unified multi-view reasoning model for 3D geometry and cameras.

%% file: sec/2_related_work.tex
\begin{figure*}[t]
\centering
\includegraphics[width=1.0\textwidth]{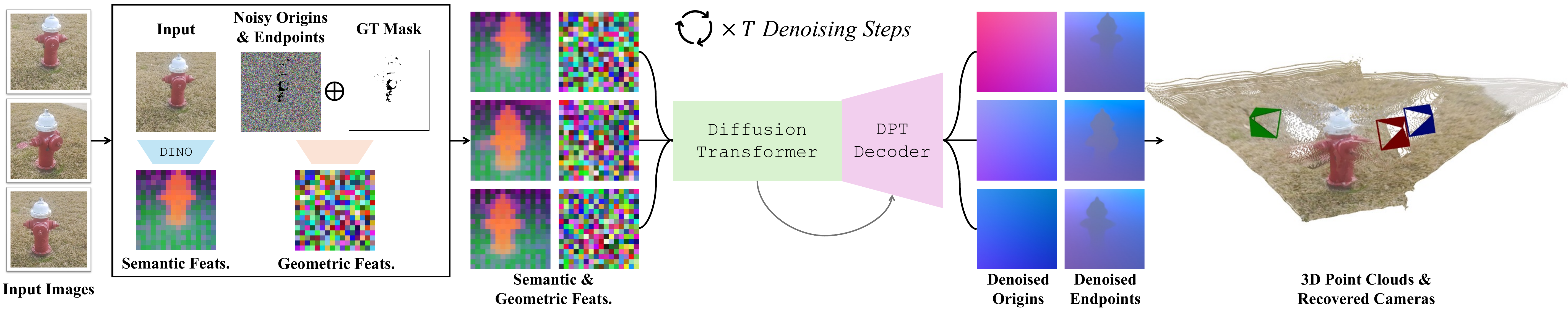}

\caption{\textbf{Method.} Given sparse multi-view images as input, \ours\ predicts pixel-wise ray origins and endpoints in a global frame (Sec. \ref{sec:origins_and_endpoints}) using a denoising diffusion process (Sec. \ref{sec:architecture}). For each image, we compute patch-wise embeddings with DINOv2~\cite{oquab2023dinov2} and embed noisy ray origins and endpoints into latents using a single downsampling convolutional layer, ensuring alignment with the spatial footprint of the image embeddings. We implement a Diffusion Transformer architecture that predicts clean ray origins and endpoints from noisy samples. A convolutional DPT \cite{ranftl2021vision} head outputs full-resolution denoised ray origins and endpoints. To handle incomplete ground truth (GT) during training, we condition the model on GT masks (Sec. \ref{sec:training_details}). At inference, the GT masks are set to all ones, enabling the model to predict origins and endpoints for all pixels. The predicted ray origins and endpoints can be directly visualized in 3D or post-processed to recover camera extrinsics, intrinsics, and multi-view consistent depth maps.}
\label{fig:framrwork}
\end{figure*}

\section{Related Work}
\label{sec:related_work}

\noindent\textbf{Structure from Motion.} Structure-from-Motion (SfM) systems \cite{schonberger2016structure} aim to simultaneously recover geometry and cameras given a set of input images. The typical SfM pipeline extracts pixel correspondences from keypoint matching \cite{lowe2004distinctive, bay2006surf}, and performs global bundle adjustment (BA) to optimize sparse 3D points and camera parameters by minimizing reprojection errors. Recently, SfM pipelines have been substantially enhanced by replacing classical subcomponents with learning-based methods, such as neural feature descriptors \cite{detone2018superpoint, edstedt2024roma}, keypoint matching \cite{sarlin2020superglue,sun2021loftr, lindenberger2023lightglue}, and bundle adjustment \cite{tang2018ba, lindenberger2021pixel}.

More recently, an emerging body of research aims to unify the various SfM subcomponents into an end-to-end neural framework. Notably, ACEZero \cite{brachmann2024scene} fits a single neural network to input images and learns pixel-aligned 3D coordinates in a self-supervised manner, while FlowMap \cite{smith2024flowmap} predicts per-frame cameras and depth maps using off-the-shelf optical flow as supervision. Though ACEZero and FlowMap are promising attempts to revolutionize SfM pipelines, they both register images incrementally and may suffer under large viewpoint changes. DUSt3R \cite{wang2023DUSt3R} directly regresses 3D pointmaps from image pairs and shows strong generalization ability \cite{li2018megadepth, reizenstein2021common}. Building on DUSt3R, MASt3R~\cite{leroy2024grounding} introduces a feature head, offering pixel-matching capabilities. MASt3R-SfM \cite{duisterhof2024mast3r} is a more scalable SfM pipeline based on MASt3R. While these approaches \cite{wang2023DUSt3R, leroy2024grounding, duisterhof2024mast3r} show impressive performance and robustness under sparse views, they are essentially pair-based, requiring sophisticated global alignment procedures to form a consistent estimate for more than two views.

\noindent\textbf{Pose Estimation with Global Reasoning.} For the task of sparse-view pose estimation, learning-based methods equipped with global reasoning show favorable robustness where traditional SfM methods \cite{snavely2006photo, schonberger2016structure} fail. This line of research includes energy-based \cite{zhang2022relpose, lin2023relpose++}, regression-based \cite{sinha2023sparsepose}, and diffusion-based pose estimators \cite{wang2023posediffusion, zhang2024cameras}. Among them, diffusion-based methods show a better ability to handle uncertainty \cite{zhang2024cameras}. Closest to our work, RayDiffusion \cite{zhang2024cameras} leverages a denoising diffusion model \cite{ho2020denoising, peebles2023scalable} with a patch-aligned (depth-unaware) ray representation to predict generic cameras. Our method goes further and pursues a generic representation for both geometry and cameras in the form of ray origins and endpoints for each pixel. In addition to resulting in a richer output, this joint geometry and pose prediction also yields improvements for pose estimation.

%% file: sec/3_method.tex
\section{Method}
\label{sec:method}

Given a set of sparse (\ie, 2-8) input images, \ours\ predicts the geometry and cameras of a 3D scene in a global coordinate frame. In Sec. \ref{sec:origins_and_endpoints}, we propose to represent 3D scenes as dense pixel-aligned ray origins and endpoints. To predict such scene representations from sparse input images while modeling uncertainty, Sec. \ref{sec:architecture} proposes a denoising diffusion architecture. We then discuss some key practical challenges in training such a model in Sec. \ref{sec:training_details}. 

\subsection{3D Scenes as Ray Origins and Endpoints}
\label{sec:origins_and_endpoints}
Given an input image $\mathbf{I} \in \mathbb{R}^{H \times W \times 3}$ with a depth map $\mathbf{D} \in \mathbb{R}^{H \times W}$, camera intrinsics $\mathbf{K} \in \mathbb{R}^{3 \times 3}$, and world-to-camera extrinsics $\mathbf{T} \in \mathbb{R}^{4 \times 4}$ (equivalently, rotation $\mathbf{R} \in SO(3)$ and translation $\mathbf{t} \in \mathbb{R}^{3}$), each 2D image pixel $\mathbf{P}_{ij} = [u, v]$ corresponds to a ray that travels from the camera center $\mathbf{c}$ through the pixel's projected position on the image plane, terminating at the object's surface as specified by the depth map $\mathbf{D}$. The endpoint of the ray associated with image pixel $\mathbf{P}_{ij}$ is given by:
\begin{equation} 
\label{eq:end_point}
\mathbf{E}_{ij} = \mathbf{T}^{-1} h \left( \mathbf{D}_{ij} \cdot \mathbf{K}^{-1} [u, v, 1]^T \right)
\end{equation}
where $h$ maps the 3D point into homogeneous coordinates. The shared ray origin $\mathbf{O}_{ij}$ for all pixels is equivalent to the camera center $\mathbf{c}$, and can be computed as:
\begin{equation} 
\label{eq:origin}
\mathbf{O}_{ij} = \mathbf{c} = h \left(-\mathbf{R}^{-1}\mathbf{t}\right)
\end{equation}
In summary, we associate each image pixel with a ray origin and endpoint $\mathbf{S}_{ij} = \langle \mathbf{O}_{ij}, \mathbf{E}_{ij} \rangle$ in world coordinates, describing the location of the observing camera and the observed 3D point on the object surface. Given a bundle of ray origins and endpoints, we can easily extract the corresponding camera pose \cite{zhang2024cameras}.

\noindent\textbf{Learning Over-Parameterized Representations.} We represent 3D scenes and cameras using distributed ray origins $\mathbf{O}$ and endpoints $\mathbf{E}$ rather than global alternatives such as quaternions and translation vectors. This design is inspired by RayDiffusion~\cite{zhang2024cameras}, and it facilitates the use of the distributed deep features learned by state-of-the-art vision backbones, such as DINOv2~\cite{oquab2023dinov2}, which encode image information in a patch-wise manner. Notably, while ray origins $\mathbf{O}$ should ideally be identical across all pixels, we predict them densely alongside ray endpoints $\mathbf{E}$. This encourages ray origins to remain close within the same image, providing implicit regularization during training. Practically, predicting both ray origins and endpoints simultaneously is easy to implement using a single projection head.

\subsection{DiffusionSfM}
\label{sec:architecture}
We propose a denoising Diffusion Transformer (DiT) architecture \cite{peebles2023scalable} that predicts ray origins and endpoints (Sec. \ref{sec:origins_and_endpoints}) via a denoising diffusion process. An overview of \ours\ is given in Fig.~\ref{fig:framrwork}.

\noindent\textbf{Diffusion Framework.} Given pixel-aligned ray origins and endpoints $\mathcal{S}=\texttt{stack}(\{\mathbf{S}^{(n)}\}_{n=1}^N)$ associated with a set of $N$ input images, we apply a forward diffusion process \cite{ho2020denoising, song2020denoising} that adds time-dependent Gaussian noise to them. Let $\mathcal{S}_t$ denote the noisy ray origins and endpoints at timestep $t$, where $\mathcal{S}_0$ is the clean sample and $\mathcal{S}_T$ (at the final diffusion step $T$) approximates pure Gaussian noise. The forward diffusion process is defined as:  
\begin{equation} 
\label{eq:forward_diffusion}
\mathcal{S}_t = \sqrt{\bar{\alpha}_t} \, \mathcal{S}_0 + \sqrt{1 - \bar{\alpha}_t} \, \epsilon
\end{equation}
where $t\sim\texttt{Uniform}(0,T]$, $\epsilon\sim\mathcal{N}(0,I)$, and $\bar{\alpha}_t$ follows a pre-defined noise schedule that controls the strength of added noise at each timestep. To perform the reverse diffusion process, which progressively reconstructs the clean sample given noisy observations, we train a diffusion model $f_\theta$ that takes $\mathcal{S}_t$ as input and optionally incorporates additional conditioning information $\mathcal{C}$. The model is trained using the following loss function (with the ``$x_0$'' objective):
\begin{equation}
\label{eq:diffusion_loss}
    \mathcal{L}_\text{Diffusion} = \mathbb{E}_{t, \mathcal{S}_0, \epsilon}
    \| \mathcal{S}_0 - f_\theta(\mathcal{S}_t, t, \mathcal{C}) \|^2
\end{equation}

\noindent\textbf{Architecture.} We implement \( f_\theta \) using a DiT \cite{peebles2023scalable} architecture conditioned on deep image features $\mathcal{C}\in\mathbb{R}^{N\times h\times w\times c_1}$ from DINOv2 \cite{oquab2023dinov2}, where $h$ and $w$ are the patch resolution and $c_1$ is the embedding dimension. To align pixels to the spatial information learned by DINOv2, we apply a convolutional layer that spatially downsamples the noisy ray origins and endpoints $\mathcal{S}_t$ to match the DINOv2 features while increasing their feature dimension:
\begin{equation}
\label{eq:line_segment_embedding}
    \mathcal{F} = \texttt{Conv}(\mathcal{S}_t) \in \mathbb{R}^{N\times h \times w \times c_2}
\end{equation}
The combined DiT input is constructed by concatenating these two feature sets along the channel dimension: $\mathcal{F}\oplus\mathcal{C}$. Within DiT, patch-wise features attend to others through self-attention \cite{vaswani2017attention}. To distinguish between different images and their respective patches, we apply sinusoidal positional encoding \cite{vaswani2017attention} based on image and patch indices.

While the DiT operates on low-resolution features, our objective is to produce pixel-aligned dense ray origins and endpoints. To achieve this, we employ a DPT (Dense Prediction Transformer) \cite{ranftl2021vision} decoder, which takes intermediate feature maps from both DINOv2 and DiT as inputs. The DPT decoder progressively increases the feature resolution through several convolutional layers. The final ray origins and endpoints are decoded from the DPT output using a single linear layer. During inference, we apply the trained model in the reverse diffusion process to iteratively denoise a randomly initialized Gaussian sample.

\begin{figure*}[t]
\centering
\includegraphics[width=0.95\textwidth]{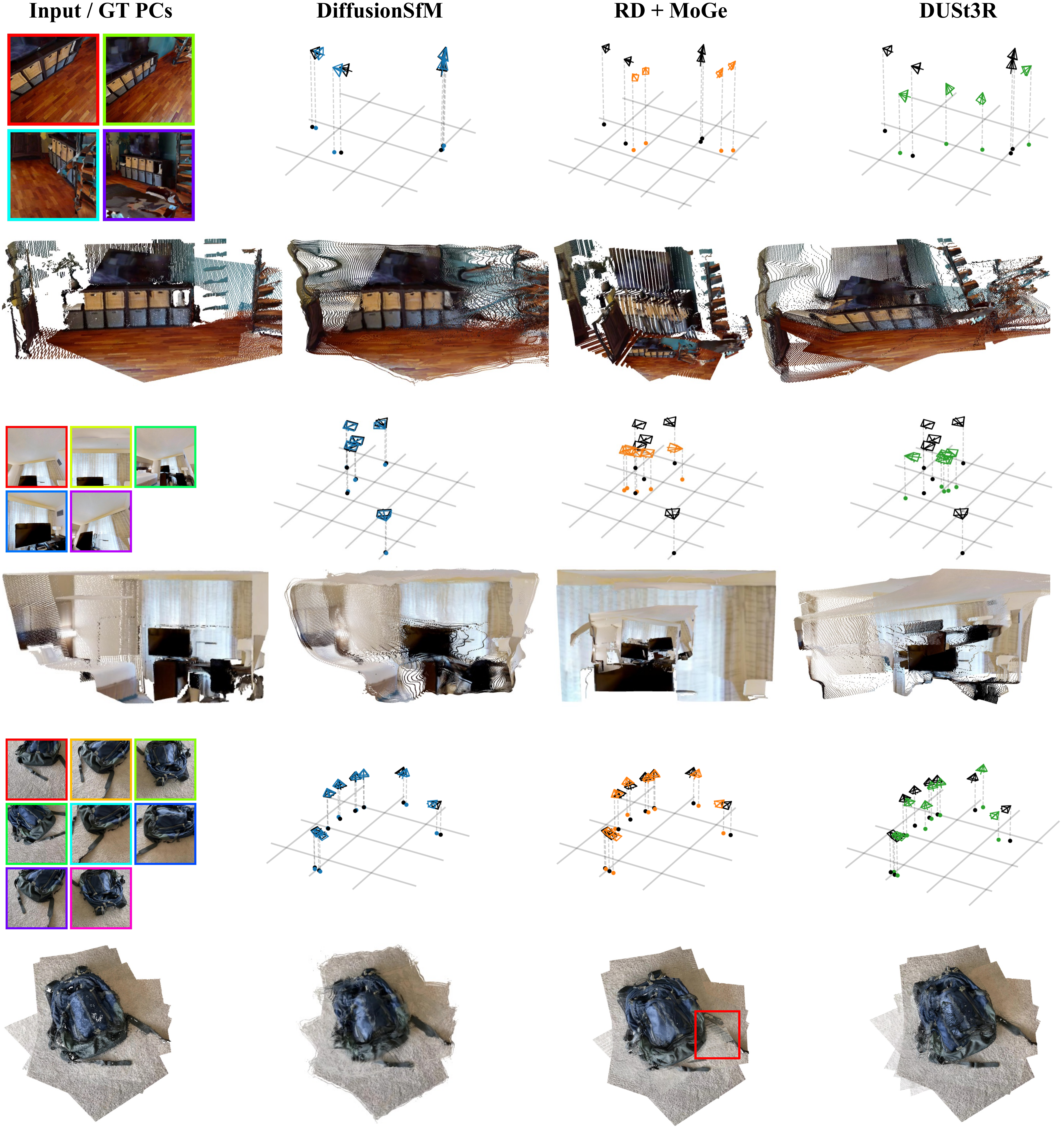}
\caption{\textbf{Qualitative Comparison on Camera Pose Accuracy and Predicted Geometry.} For each method, we plot the ground-truth cameras in black and the predicted cameras in other colors. \ours\ demonstrates robust performance even with challenging inputs. Compared to DUSt3R, which sometimes fails to register images in a consistent manner, \ours\ consistently yields a coherent global prediction. Additionally, while we observe that DUSt3R can predict highly precise camera rotations, it often struggles with camera centers (see the backpack example). Input images depicting scenes are out-of-distribution for RayDiffusion, as it is trained on CO3D only.} 
\label{fig:vis_compare}
\vspace{-1em}
\end{figure*}

\begin{figure*}[t]
\centering
\includegraphics[width=0.95\textwidth]{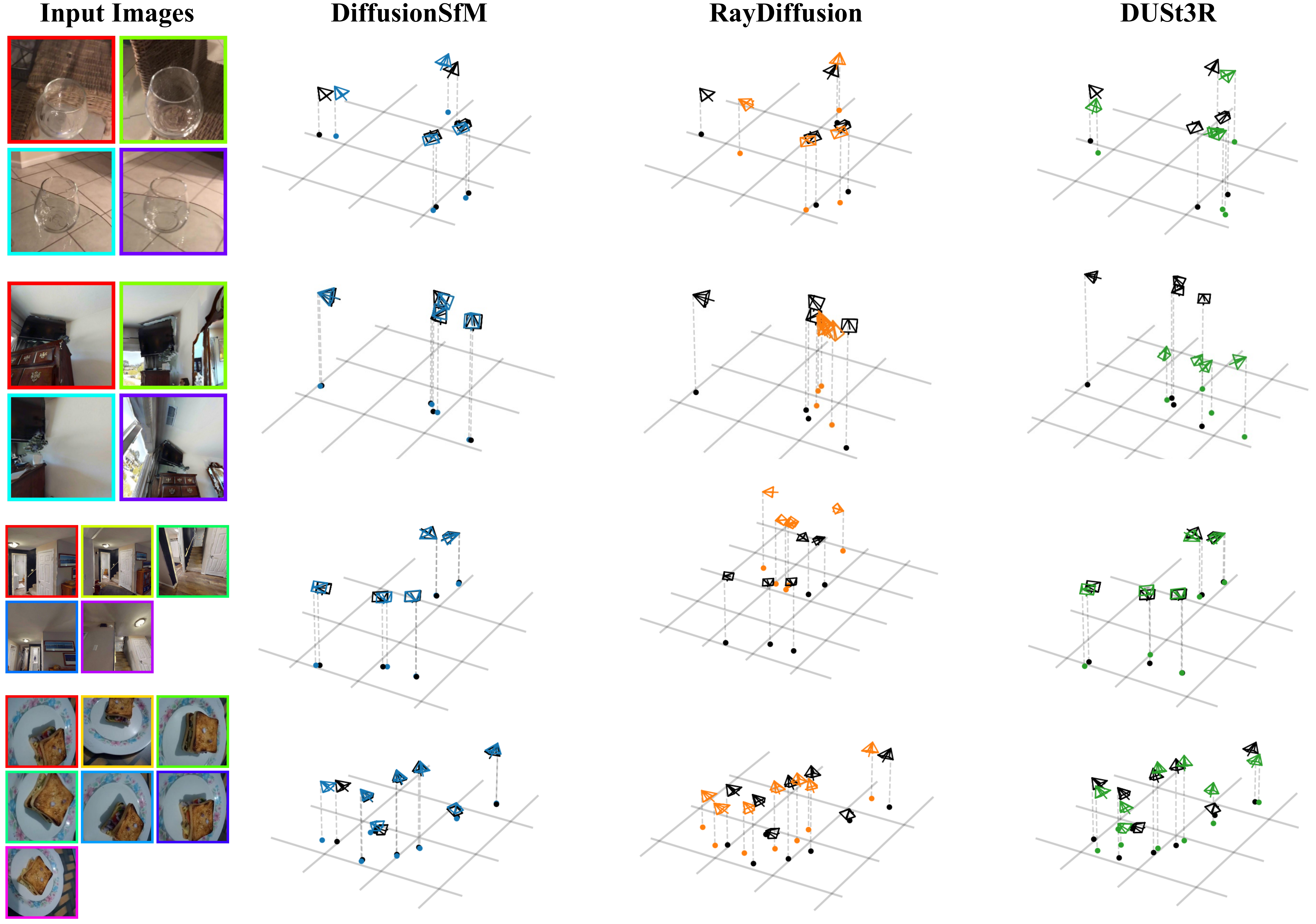}
\caption{\textbf{Additional Qualitative Results on Predicted Camera Poses.} \ours\ shows robustness to ambiguous patterns in inputs.}
\label{fig:vis_cameras}
\vspace{-10pt}
\end{figure*}

\subsection{Practical Training Considerations}
\label{sec:training_details}
\noindent\textbf{Homogeneous Coordinates for Unbounded Geometry.} 
Real-world scenes often exhibit significant variations in scale, both across different scenes and within a single scene. While normalizing by the average distance of 3D points \cite{wang2023DUSt3R} helps address cross-scene variation, within-scene variations remain. For instance, a background building may be much farther away than foreground elements, potentially resulting in extremely large coordinate values when generating ground-truth ray origins and endpoints. However, neural networks (especially diffusion models) tend to train most effectively when working with bounded inputs and outputs (\eg, between -1 and 1). To stabilize training across the large scale variations present in 3D scene datasets, we propose to represent ray origins and endpoints in homogeneous coordinates. Specifically, given any 3D point, we apply a homogeneous transform from $\mathbb{R}^3\to\mathbb{P}^3$:
\begin{equation}
    (x,y,z)\to \frac1w(x,y,z,1)
\end{equation}
where $w$ is an arbitrary scale factor. To encourage bounded coordinates in practice, we choose $w$ such that the homogeneous coordinate is unit-norm:
\begin{equation}
    w:=\sqrt{x^2+y^2+z^2+1}
\end{equation}
Unit normalization allows homogeneous coordinates to serve as a bounded representation for unbounded scene geometry. For example, $(x,y,z,0)$ is a point at infinity in the direction of $(x,y,z)$. We find that this representation makes large coordinate values more tractable during training.

\noindent\textbf{Training with Incomplete Ground Truth.} Many real-world datasets (\eg, CO3D~\cite{reizenstein2021common} and MegaDepth~\cite{li2018megadepth}), provide only sparse point clouds, leading to incomplete depth information. This presents a significant challenge in diffusion training, as ground-truth (GT) depth values that are used to create clean ray endpoints often contain invalid or missing data. It is highly undesirable for these missing ray endpoints to be interpreted as part of the target distribution. Unlike regression models (\eg, DUSt3R \cite{wang2023DUSt3R}), which do not require (noisy) GT data as \textit{input} and can simply apply masks on the loss for supervision, diffusion models must handle incomplete inputs during training.

To mitigate this issue, we further apply GT masks $\mathcal{M}\in\mathbb{R}^{N\times H\times W}$ to the DiT inputs, where zero values indicate pixels with invalid depth. During training, we multiply noisy rays with GT masks element-wise, then concatenate along the channel dimension: $\mathcal{S}'_t=(\mathcal{M}\cdot\mathcal{S}_t)\oplus\mathcal{M}$. Then, we only compute the diffusion loss in Eq.~\ref{eq:diffusion_loss} over unmasked pixels. By implementing these strategies, we encourage the model to focus on regions with valid GT values during training. During inference, however, we would like the diffusion process to estimate ray origins and endpoints at all pixels, so we always use GT masks with values set to one.

\noindent\textbf{Sparse-to-Dense Training.} In practice, we find that training the entire model from scratch leads to slow convergence and suboptimal performance. To address this, we propose a sparse-to-dense training approach. First, we train a sparse version of the model, where the DPT decoder is removed, and the output ray origins and endpoints have the same spatial resolution as the DINOv2 features. Unlike Eq.~\ref{eq:line_segment_embedding}, no spatial downsampling is required, so this sparse model uses a single linear layer to embed the noisy ray origins and endpoints. Once the sparse model is trained, we initialize the dense model DiT with the learned weights from the sparse model. This two-stage approach significantly improves performance; see the supplementary material for comparisons.

%% file: sec/4_experiments.tex
\section{Experiments}

\begin{table*}[t]
    \small{
    \centering
    \begin{tabular}{ll|ccccccc|ccccccc}
\toprule
& & \multicolumn{7}{c}{Rotation Accuracy ($\uparrow$, @ 15$^\circ$)}
  & \multicolumn{7}{c}{Center Accuracy ($\uparrow$, @ 0.1)}\\
& \# of Images & 2 & 3 & 4 & 5 & 6 & 7 & 8
  & 2 & 3 & 4 & 5 & 6 & 7 & 8 \\
\midrule
\parbox[t]{2mm}{\multirow{8}{*}{\rotatebox[origin=c]{90}{Seen Categories}}} 
& COLMAP~\cite{sarlin2019coarse}
  & 30.7 &  28.4 &  26.5 &  26.8 &  27.0 &  28.1 &  30.6
  & 100 & 34.5 & 23.8 & 18.9 & 15.6 & 14.5 & 15.0 \\
& PoseDiffusion~\cite{wang2023posediffusion}
  & 75.7 & 76.4 & 76.8 & 77.4 & 78.0 & 78.7 & 78.8
  & 100 & 77.5 & 69.7 & 65.9 & 63.7 & 62.8 & 61.9 \\
& RelPose++~\cite{lin2023relpose++}
  & 81.8 & 82.8 & 84.1 & 84.7 & 84.9 & 85.3 & 85.5
  & 100 &  85.0 & 78.0 & 74.2 & 71.9 & 70.3 & 68.8 \\
& RayDiffusion~\cite{zhang2024cameras}
  & 91.8 & 92.4 & 92.6 & 92.9 & 93.1 & 93.3 & 93.3
  & 100 & 94.2 & 90.5 & 87.8 & 86.2 & 85.0 & 84.1 \\
& DUSt3R-CO3D~\cite{wang2023DUSt3R}
  & 86.7 & 87.9 & 88.0 & 88.2 & 88.6 & 88.8 & 88.9
  & 100 & 92.0 & 86.8 & 83.8 & 82.0 & 81.1 & 80.4 \\
& DUSt3R~\cite{wang2023DUSt3R}
  & 91.7 & 92.7 & 93.3 & 93.6 & 93.8 & 94.0 & 94.3
  & 100 & 93.0 & 85.7 & 81.9 & 79.6 & 77.8 & 76.8 \\
& DiffusionSfM-CO3D
  & \textbf{93.4} & \underline{94.0} & \underline{94.5} & \underline{94.8} & \underline{95.0} & \underline{95.2} & \underline{95.1}
  & 100 & \textbf{95.9} & \textbf{93.6} & \underline{92.2} & \underline{91.2} & \underline{90.7} & \underline{90.2} \\
& DiffusionSfM
  & \underline{92.6} & \textbf{94.1} & \textbf{94.6} & \textbf{95.0} & \textbf{95.3} & \textbf{95.5} & \textbf{95.5}
  & 100 & \underline{95.6} & \underline{93.4} & \textbf{92.4} & \textbf{91.7} & \textbf{91.1} & \textbf{90.7} \\
\midrule
\parbox[t]{2mm}{\multirow{8}{*}{\rotatebox[origin=c]{90}{Unseen Categories}}}
& COLMAP~\cite{sarlin2019coarse}
  & 34.5 &  31.8 &  31.0 &  31.7 &  32.7 &  35.0 &  38.5
  & 100 & 36.0 & 25.5 & 20.0 & 17.9 & 17.6 & 19.1 \\
& PoseDiffusion~\cite{wang2023posediffusion}
  & 63.2 & 64.2 & 64.2 & 65.7 & 66.2 & 67.0 & 67.7
  & 100 & 63.6 & 50.5 & 45.7 & 43.0 & 41.2 & 39.9 \\
& RelPose++~\cite{lin2023relpose++}
  & 69.8 & 71.1 & 71.9 & 72.8 & 73.8 & 74.4 & 74.9
  & 100 & 70.6 & 58.8 & 53.4 & 50.4 & 47.8 & 46.6 \\
& RayDiffusion~\cite{zhang2024cameras}
  & 83.5 & 85.6 & 86.3 & 86.9 & 87.2 & 87.5 & 88.1
  & 100  & 87.7 & 81.1 & 77.0 & 74.1 & 72.4 & 71.4 \\
& DUSt3R-CO3D~\cite{wang2023DUSt3R}
  & 79.8 & 81.5 & 82.6 & 82.7 & 83.0 & 83.3 & 83.7
  & 100 & 83.6 & 77.2 & 71.8 & 70.0 & 68.1 & 67.0 \\
& DUSt3R~\cite{wang2023DUSt3R}
  & \underline{90.8} & \underline{92.6} & \underline{93.6} & \underline{93.6} & \underline{93.8} & \underline{93.6} & 93.4
  & 100 & 87.9 & 79.8 & 74.3 & 71.7 & 69.4 & 67.8 \\
& DiffusionSfM-CO3D
  & 90.4 & 91.2 & 92.7 & 93.0 & 93.1 & 93.3 & \underline{93.5}
  & 100 & \underline{91.1} & \underline{87.7} & \underline{85.3} & \underline{83.7} & \underline{82.7} & \underline{82.0} \\
& DiffusionSfM
  & \textbf{91.3} & \textbf{92.8} & \textbf{93.8} & \textbf{94.5} & \textbf{95.0} & \textbf{95.1} & \textbf{95.3}
  & 100 & \textbf{92.6} & \textbf{88.4} & \textbf{87.0} & \textbf{86.4} & \textbf{85.1} & \textbf{84.7} \\
\bottomrule
\end{tabular}
}
\caption{
    \textbf{Camera Rotation and Center Accuracy on CO3D.} On the left, we report the proportion of relative camera rotations within $15^\circ$ of the ground truth. On the right, we report the proportion of camera centers within $10\%$ of the scene scale, relative to the ground truth. To align the predicted camera centers to ground truth, we apply an optimal similarity transform ($s$, $\mathbf{R}$, $\mathbf{t}$), hence the alignment is perfect at $N=2$ but worsens with more images. DiffusionSfM outperforms all other methods for camera center accuracy, and outperforms all methods trained on equivalent data for rotation accuracy.
    }
    \vspace{-2mm}
    \label{tab:co3d_rot_tra}
\end{table*}

\subsection{Experimental Setup}
\noindent\textbf{Datasets.} We introduce two model variants, each trained on different datasets. (1) DiffusionSfM-CO3D: Following prior work \cite{lin2023relpose++, zhang2024cameras}, we train and evaluate our model on the CO3D dataset \cite{reizenstein2021common}, which consists of turntable video sequences of various object categories. Specifically, we train on 41 object categories and evaluate on both these \textit{seen} categories and an additional 10 \textit{unseen} categories to assess generalization. (2) DiffusionSfM: This variant is trained on the datasets used for DUSt3R \cite{wang2023DUSt3R}, excluding Waymo \cite{sun2020scalability} due to its excessively sparse depth maps. The included datasets are Habitat \cite{savva2019habitat}, CO3D \cite{reizenstein2021common}, ScanNet++ \cite{yeshwanth2023scannet++}, ArkitScenes \cite{baruch2021arkitscenes}, Static Scenes 3D \cite{mayer2016large}, MegaDepth \cite{li2018megadepth}, and BlendedMVS \cite{yao2020blendedmvs}. We follow the DUSt3R official repository \cite{wang2023DUSt3RRepo} guidelines to extract image pairs, ensuring reasonable overlap between paired images. To construct multi-view data given these pre-computed pairs, we iteratively sample images using an adjacency matrix, maintaining sufficient overlap with the selected set. Beyond CO3D, this model variant is also evaluated on Habitat and RealEstate10k \cite{zhou2018stereo}.

\noindent\textbf{Baselines and Metrics.} To evaluate camera pose accuracy in the sparse-view setup, we compare with RayDiffusion \cite{zhang2024cameras} and DUSt3R \cite{wang2023DUSt3R}, along with previous methods \cite{sarlin2019coarse, wang2023posediffusion, lin2023relpose++}. DUSt3R is trained initially on a mixture of eight datasets, while most other baselines are trained only on CO3D. For a fair and comprehensive comparison, we re-train DUSt3R on the 41-10 split of CO3D, using the authors' official implementation \cite{wang2023DUSt3RRepo} and hyperparameters (referred to as DUSt3R-CO3D). To evaluate camera predictions, we follow prior work \cite{zhang2024cameras} and convert predicted rays back to traditional extrinsic matrices and report two pose accuracy metrics: (1) Camera Rotation Accuracy which compares the predicted relative camera rotation between images against ground truth and (2) Camera Center Accuracy which compares predicted camera centers to the ground truth after a similarity alignment. To evaluate the estimated geometry (\ie, ray endpoints), we report Chamfer Distance (CD). Additionally, to explore whether combining an off-the-shelf sparse-view pose estimation method (RayDiffusion) with a monocular depth estimation model (MoGe \cite{wang2024moge}) is sufficient to infer 3D scene geometry from multiple images, we introduce a new baseline, $\texttt{RD+MoGe}$. We include more details in the supplementary. Unless otherwise specified, all evaluations use 2–8 input images.

\subsection{Evaluation on CO3D}
\noindent\textbf{Camera Pose Accuracy.} We present the quantitative results in Tab.~\ref{tab:co3d_rot_tra}. For camera rotation accuracy, both versions of \ours\ outperform all baselines, except that DUSt3R achieves slightly higher overall accuracy than \ours-CO3D on unseen categories -- likely due to its training on more data. For camera center accuracy, our approach consistently outperforms all other methods. We hypothesize that these gains stem from our explicit modeling of camera centers through ray origin prediction. The qualitative results in Fig.~\ref{fig:vis_cameras} illustrate that \ours\ produces robust predictions given challenging inputs, whereas DUSt3R produces inaccurate results. We attribute this improvement to our model’s probabilistic modeling capability derived from diffusion, as well as its multi-view reasoning abilities, which together effectively handle these challenging scenarios. While we observe that DUSt3R can predict highly precise camera rotations, it often struggles with camera centers (see the backpack example in Fig.~\ref{fig:vis_compare}).

\noindent\textbf{Predicted Geometry.} To evaluate predicted geometry, we compute Chamfer Distance (CD) and show comparisons against baselines in Tab.~\ref{tab:co3d_cd}. We compute CD in two setups (with and without foreground object masks), and find that \ours-CO3D performs best without foreground masking. In this setup, the predicted ray endpoints corresponding to the background image pixels tend to have larger coordinate values than foreground ones, and therefore dominate CD. This result indicates that our model provides more accurate predictions for complex image backgrounds. In terms of CD with masking, DUSt3R achieves the best result, while our two model variants outperform DUSt3R-CO3D. See Fig.~\ref{fig:vis_compare} for visualization. We also notice that naively combining RayDiffusion poses with estimated depths from MoGe does not work well, as the inferred depths are potentially inconsistent across different views, thus resulting in degraded performance.

\begin{table}[t]
    \centering
    \small{
    \setlength{\tabcolsep}{2pt}
    \begin{tabular}{lccccccc}
\toprule
\# of Images & 2 & 3 & 4 & 5 & 6 & 7 & 8\\
\midrule
RD*+MoGe~\cite{zhang2024cameras} & 0.059 & 0.064 & 0.071 & 0.062 & 0.063 & 0.061 & 0.061 \\
DUSt3R*~\cite{wang2023DUSt3R}
& 0.036 & 0.037 & 0.040 & 0.040 & 0.037 & 0.036 & 0.039 \\
DUSt3R~\cite{wang2023DUSt3R}
  & \underline{0.021} & \underline{0.023} & \underline{0.024} & \underline{0.024} & \underline{0.025} & \underline{0.025} & \underline{0.023} \\
\ours* 
  & \textbf{0.019} & \textbf{0.021} & \textbf{0.022} & \textbf{0.022} & \textbf{0.021} & \textbf{0.021} & \textbf{0.021} \\
\ours & 0.024 & 0.024 & 0.025 & \underline{0.024} & \underline{0.025} & 0.026 & 0.027 \\
\midrule
RD*+MoGe~\cite{zhang2024cameras} & 0.071 & 0.075 & 0.068 & 0.067 & 0.066 & 0.064 & 0.064 \\
DUSt3R*~\cite{wang2023DUSt3R}
  & 0.038 & 0.036 & 0.036 & 0.036 & 0.034 & 0.033 & 0.034 \\
DUSt3R~\cite{wang2023DUSt3R}
  & \textbf{0.023} & \textbf{0.022} & \textbf{0.019} & \textbf{0.020} & \textbf{0.019} & \textbf{0.020} & \textbf{0.020} \\
\ours*
  & \underline{0.028} & 0.025 & 0.024 & 0.024 & 0.024 & 0.023 & \underline{0.022} \\
\ours
  & \underline{0.028} & \underline{0.023} & \underline{0.022} & \underline{0.022} & \underline{0.023} & \underline{0.021} & \textbf{0.020} \\
\bottomrule
\end{tabular}
}
\caption{\textbf{Chamfer Distance ($\downarrow$) on CO3D Unseen Categories.} \textbf{Top:} CD computed on all scene points. \textbf{Bottom:} CD computed on foreground points only. Models marked with ``*'' are trained on CO3D only, while those without are trained on multiple datasets. Note that top and bottom values are not directly comparable, as each ground-truth point cloud is individually normalized.}
\label{tab:co3d_cd}
\end{table}

\begin{table}[t]
\centering
\small
\resizebox{\linewidth}{!}{
    \begin{tabular}{clcccc}
        \toprule
        \parbox[t]{2mm}{\multirow{2}{*}{\rotatebox[origin=c]{90}{\footnotesize Hab.}}}
            & DUSt3R~\cite{wang2023DUSt3R} & \textbf{97.0}/100 & \textbf{95.0}/97.6 & 94.3/95.0 & 94.2/93.1 \\
            & \ours & 92.7/100 & 93.9/\textbf{99.0} & 94.3/\textbf{98.6} & \textbf{94.7}/\textbf{98.4} \\
        \midrule
        \parbox[t]{2mm}{\multirow{2}{*}{\rotatebox[origin=c]{90}{\scriptsize RE10}}} 
            & DUSt3R~\cite{wang2023DUSt3R} & \textbf{98.1}/100 & 97.7/68.7 & 97.6/57.9 & 97.7/53.3 \\
            & \ours & 97.9/100 & \textbf{97.8}/\textbf{74.9} & \textbf{98.0}/\textbf{67.7} & \textbf{98.0}/\textbf{63.7} \\
        \bottomrule
    \end{tabular}
}
\caption{\textbf{Camera Rotation and Center Accuracy on Two Scene-Level Datasets.} \textbf{Top:} Habitat (2–5 views). \textbf{Bottom:} RealEstate10k (2, 4, 6, 8 views). Each grid reports camera rotation accuracy (left, $\uparrow$) and center accuracy (right, $\uparrow$). While \ours\ performs on par with DUSt3R in rotation accuracy, it consistently surpasses DUSt3R in center accuracy.}
\label{tab:eval_scene_dataset}
\end{table}

\subsection{Evaluation on Scene-Level Datasets}  
Beyond the object-centric CO3D dataset~\cite{reizenstein2021common}, we compare \ours\ against DUSt3R on two scene-level datasets: Habitat (in-distribution)~\cite{savva2019habitat} and RealEstate10k (out-of-distribution)~\cite{zhou2018stereo}. The results are presented in Tab.~\ref{tab:eval_scene_dataset}. While \ours\ achieves comparable camera rotation accuracy to DUSt3R, it consistently predicts more accurate camera centers, aligning with our findings on CO3D.

\subsection{Inference Speed}
Though our method requires iterative diffusion denoising at inference, we can speed this up by performing early stopping. Specifically, we can treat the $x_0$-prediction from early timesteps as our output instead of iterating over all denoising timesteps. Consistent with observations by Zhang \etal~\cite{zhang2024cameras}, this in fact yields more accurate predictions than the final-step diffusion outputs. As a result we only require 10 denoising diffusion timesteps for inference, taking 1.91 seconds on a single A5000 GPU with 8 input images. In contrast, DUSt3R takes 8.73 seconds to run the complete pairwise inference and global alignment procedure. We provide additional analysis in the supplementary material.

\subsection{Ablation Study}
\begin{table}[t]
    \small{
    \setlength{\tabcolsep}{2.5pt}
    \begin{tabular}{lccccccc}
\toprule
\# of Images & 2 & 3 & 4 & 5 & 6 & 7 & 8\\
\midrule
\ours*
  & \textbf{0.019} & \textbf{0.021} & \textbf{0.022} & \textbf{0.022} & \textbf{0.021} & \textbf{0.021} & \textbf{0.021} \\
w/o Mask
  & 0.020 & 0.022 & 0.023 & 0.023 & 0.022 & 0.022 & 0.024 \\
\bottomrule
\end{tabular}
}
\caption{\textbf{Ablation Study on GT Mask Conditioning for CO3D Unseen Categories.} We assess the effect of replacing GT mask conditioning with depth interpolation in our CO3D variant (\ours*), by reporting the CD for predicted geometry. Incorporating mask conditioning to indicate missing data during training improves geometry quality.}
\label{tab:ablation_mask}
\vspace{-5pt}
\end{table}

\noindent\textbf{Homogeneous Coordinates.} The use of homogeneous coordinates for ray origins and endpoints is crucial for stable model training. To assess its impact, we replace the proposed homogeneous representation with standard 3D coordinates in \(\mathbb{R}^3\). Our experiments show that this variant is difficult to train and fails to converge. Further details on this experiment are provided in the supplementary material.

\noindent\textbf{GT Mask Conditioning.} To evaluate the effectiveness of the proposed GT mask conditioning, we train a baseline model without using this strategy. For missing values in the ground-truth depth maps, we use nearest-neighbor interpolation to fill in invalid pixels. This experiment is conducted on the CO3D dataset, with results presented in Tab.~\ref{tab:ablation_mask}. The findings show that removing GT mask conditioning consistently degrades predicted geometry across varying numbers of input views, even when the loss is still masked. While interpolation can effectively fill missing depth within object regions, it often introduces substantial noise in the background (\eg, the sky). This noise negatively impacts diffusion model training, as the entire ray origin and endpoint maps are used as input. We anticipate an even larger performance gap on non-object-centric datasets with more outdoor scenes, such as MegaDepth \cite{li2018megadepth}.

\subsection{Multi-modality from Multiple Sampling}  

\begin{figure}[t]
\centering
\includegraphics[width=\linewidth]{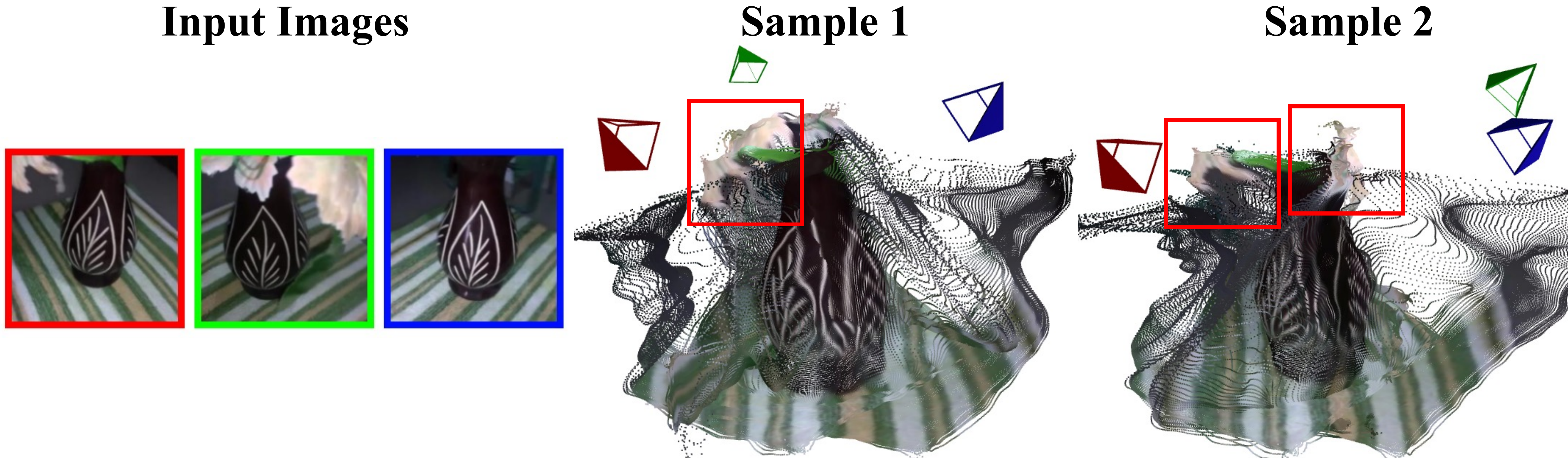}
\caption{\textbf{Multi-modality of \ours.} We show two distinct samples from \ours, starting from the same input images but with different random noise. Sample 1 explains the input images by putting all flowers on the left side, while Sample 2 places one flower on each side (note the difference in the green camera's viewpoint). \ours~is able to predict multi-modal geometry distributions when the scene layout is ambiguous in the inputs.}
\label{fig:multi_modality}
\vspace{-5pt}
\end{figure}

Diffusion models enable the generation of diverse samples from challenging input images. For instance, in Fig.~\ref{fig:multi_modality}, the vase exhibits symmetric patterns, and we present two distinct predicted endpoints from \ours, each offering a different interpretation of the flowers in the images. Compared to regression models such as DUSt3R \cite{edstedt2024roma}, \ours\ is better suited for handling uncertainty -- an inherent aspect of our task -- where a sparse set of input images can correspond to multiple plausible 3D scene geometries.

%% file: sec/5_discussion.tex
\section{Discussion}
We present \ours\ and demonstrate that it recovers accurate predictions of both cameras and geometry from multi-view inputs. Although our results are promising, several challenges and open questions remain. 

Notably, \ours\ employs a pixel-space diffusion model, in contrast to the latent-space models adopted by state-of-the-art T2I generative systems. Operating in pixel space may require greater model capacity, yet our current model remains relatively small -- potentially explaining the noisy patterns observed along object boundaries. Learning an expressive latent space for ray origins and endpoints by training a VAE could be a promising direction for future work. In addition, the computational requirement in multi-view transformers scales quadratically with the number of input images: one would require masked attention to deploy systems like ours for a large set of input images. 

Despite these challenges, we believe that our work highlights the potential of a unified approach for multi-view geometry tasks. We envision that our approach can be built upon to train a common system across related geometric tasks, such as SfM (input images with unknown origins and endpoints), registration (some images have known origins and endpoints, whereas others don't), mapping (known rays but unknown endpoints), and view synthesis (unknown pixel values for known rays).

%% file: sec/X_suppl.tex
\clearpage
\maketitlesupplementary

\appendix

\section*{Overview}
The supplementary material includes sections as follows:
\begin{itemize}
\item Section \ref{sec:supp_inference_details}: Implementation details.
\item Section \ref{sec:supp_rd_plus_moge}: Additional analysis on integrating RayDiffusion \cite{zhang2024cameras} with MoGe \cite{wang2024moge}.
\item Section \ref{sec:supp_qualitative_results}: More qualitative comparisons of predicted geometry and camera poses against baseline methods.
\item Section \ref{sec:supp_sparse_to_dense}: Details and evaluation of the sparse-to-dense training strategy employed in \ours.
\item Section \ref{sec:supp_homogeneous_representation}: More analysis of the homogeneous representation.
\item Section \ref{sec:supp_rays_to_cameras}: Converting predicted ray origins and endpoints into camera poses.
\end{itemize}

\begin{figure*}[!t]
\centering
\includegraphics[width=\textwidth]{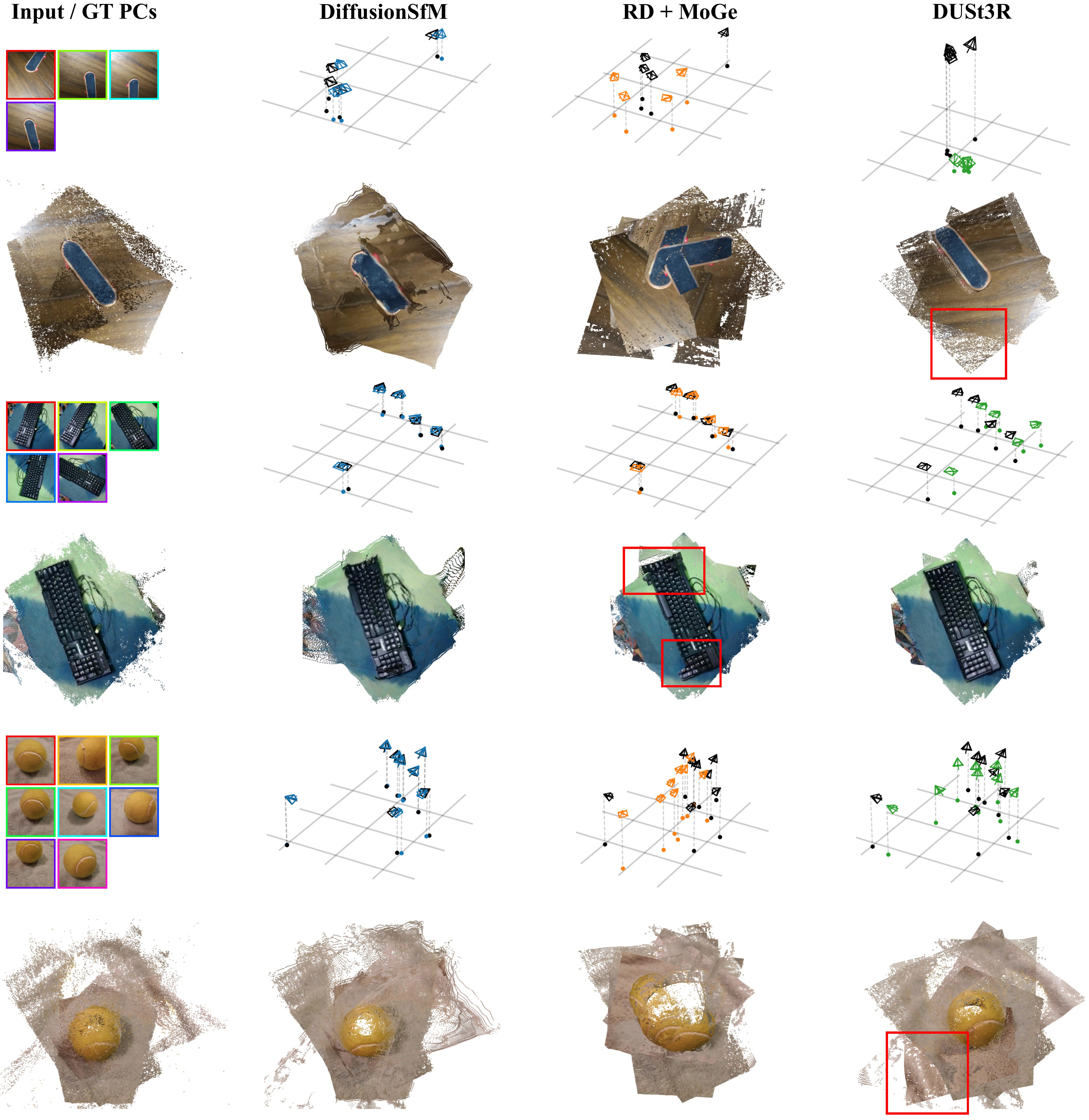}
\caption{\textbf{More Qualitative Comparisons on Predicted Geometry and Camera Poses.} DiffusionSfM shows superior capabilities in handling challenging samples, \eg, the skateboard and tennis ball. Additionally, while we observe that DUSt3R can predict highly precise camera rotations, it often struggles with camera centers (see the keyboard example).}
\label{fig:supp_vis_compare}
\end{figure*}

\begin{figure*}[!t]
\centering
\includegraphics[width=\textwidth]{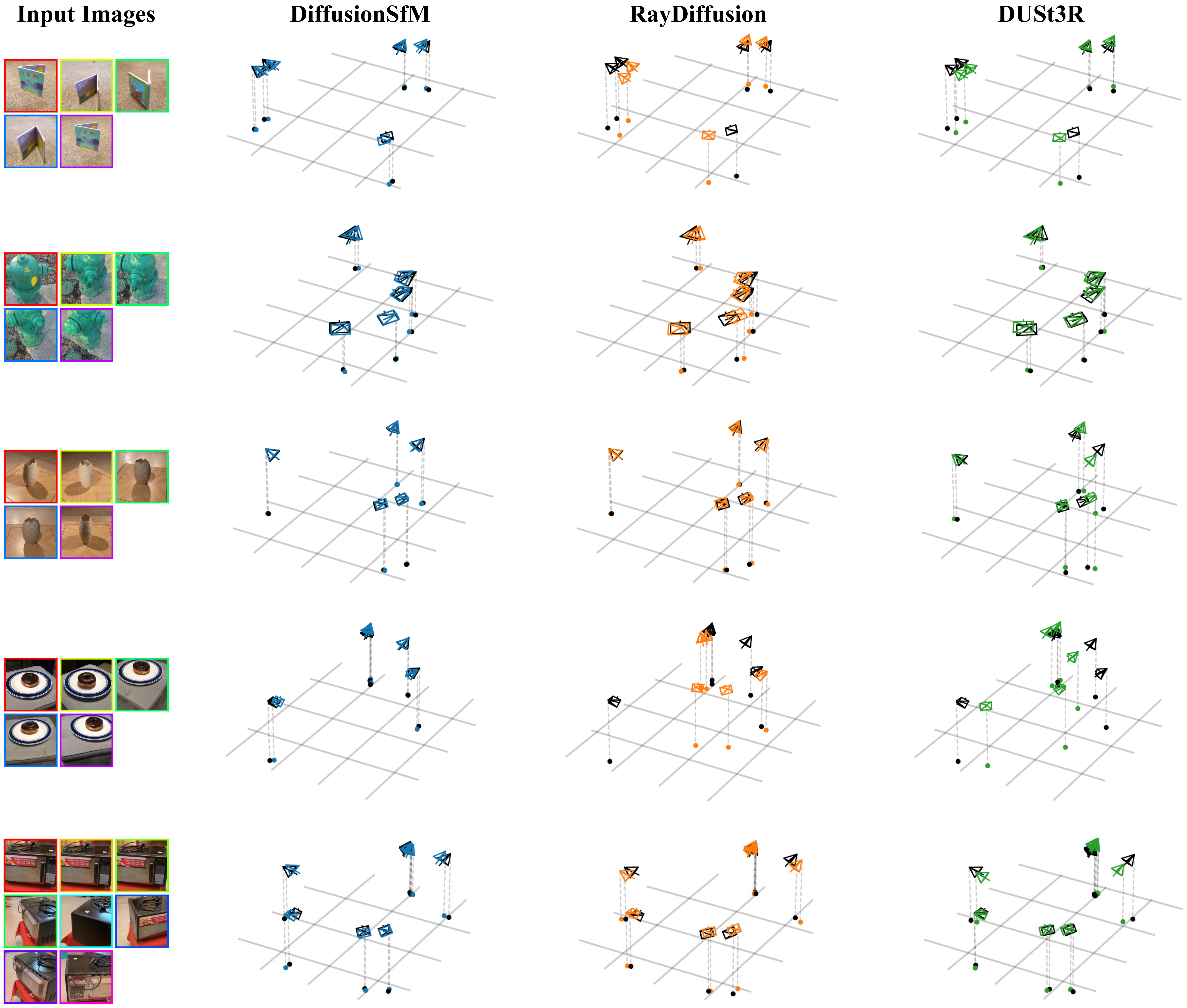}
\caption{\textbf{More Qualitative Comparisons on Predicted Camera Poses.}}
\label{fig:supp_vis_cameras}
\end{figure*}

\section{Implementation Details}
\label{sec:supp_inference_details}
\begin{figure*}[!t]
\centering
\includegraphics[width=\linewidth]{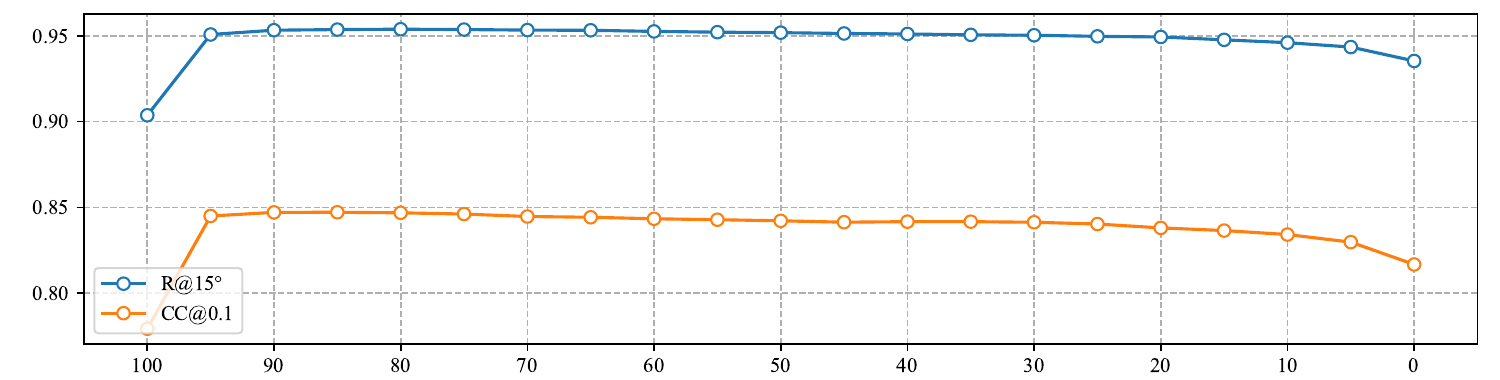}
\caption{\textbf{Performance of $x_0$-Prediction on CO3D Unseen Categories across Diffusion Denoising Timesteps (N = 8).} The X-axis represents the diffusion denoising timesteps, with $T=100$ indicating predictions starting from Gaussian noise and $T=0$ corresponding to the clean sample. The Y-axis shows the accuracy for camera rotation (blue) and camera center (orange). Notably, DiffusionSfM achieves peak performance at $T=90$. As a result, in inference, we perform only 10 diffusion steps, significantly improving inference speed.}

\label{fig:supp_early_stop}
\end{figure*}

\textbf{Inference}. DiffusionSfM utilizes $x_0$-parameterization to predict clean ray origin and endpoint maps as the model output, employing 100 diffusion denoising timesteps. In Fig.~\ref{fig:supp_early_stop}, we evaluate the accuracy of $x_0$-prediction at each timestep with eight input images on CO3D \cite{reizenstein2021common} unseen categories. Interestingly, we find that DiffusionSfM achieves its most accurate clean sample predictions at an early timestep ($T=90$), rather than at the final denoising step. This observation remains consistent across different numbers of input images (Zhang \etal~\cite{zhang2024cameras} also have a similar observation that early stopping helps improve performance). To capitalize on this property, we limit inference to 10 denoising steps and use the $x_0$-prediction at $T=90$ as the final output, significantly reducing inference time. Moreover, we find that the optimal timestep varies across datasets: $T=85$ yields the best results on Habitat \cite{savva2019habitat}, while $T=95$ performs best on RealEstate10k \cite{zhou2018stereo}. All experiments use a zero-terminal-SNR noise schedule \cite{lin2024common}.

\noindent\textbf{Resolving Ambiguities in Ground Truth.} We transform camera poses so the first camera has an identity rotation and is positioned at the world origin. For scale, we unproject the first image in the input views using ground-truth (GT) depth and scale the world coordinates so the ``median point'' lies at a unit distance from the origin. Our model is trained to conform to this scene configuration.

\section{$\texttt{RD+MoGe}$ Baseline: More Details}
\label{sec:supp_rd_plus_moge}
To minimize the scale difference for the predicted camera poses from RayDiffusion \cite{zhang2024cameras} and depths from MoGe \cite{wang2024moge} to form a single consistent output, we follow these procedures: (1) We match the MoGe depth with the GT depth using a 1D optimal alignment (thus giving this baseline some privileged information). (2) We align the predicted camera centers from RayDiffusion with GT cameras using an optimal similarity transform. (3) Finally, we unproject image pixels using the updated camera parameters and the aligned depths. We find that a naive combination of RayDiffusion and MoGe yields poor Chamfer Distance, even though RayDiffusion estimates relatively accurate focal length. This is because the MoGe depth estimates for different input views are inconsistent with each other. Therefore, to predict consistent 3D geometry from multiple images, the model must learn to reason over the entire set of views, rather than relying on mono-depth predictions from individual images. We also include more visualizations in Fig.~\ref{fig:supp_vis_compare}, where duplicated structures are observed due to significant pose errors or a minor misalignment between views.

\section{More Qualitative Comparisons}
\label{sec:supp_qualitative_results}

We include more qualitative comparisons with baselines on the predicted geometry (Fig.~\ref{fig:supp_vis_compare}) and camera poses (Fig.~\ref{fig:supp_vis_cameras}).

\noindent \textbf{Discussion.} We show that DiffusionSfM can handle challenging input images where objects present highly symmetric patterns (\eg, the tennis ball example in Fig.~\ref{fig:supp_vis_compare} and the donut example in Fig.~\ref{fig:supp_vis_cameras}), while RayDiffusion \cite{zhang2024cameras} and DUSt3R \cite{wang2023DUSt3R} fail to predict correct camera poses. Compared to RayDiffusion, our approach leverages the prediction of \textit{dense} scene geometry (\ie, pixel-aligned ray origins and endpoints) rather than relying on patch-wise ``depth-agnostic'' rays. We find that predicting dense pixel-aligned outputs improves performance (see Sec. \ref{sec:supp_sparse_to_dense}). When compared to DUSt3R, our model benefits from attending to all input images simultaneously and utilizing a diffusion framework to effectively manage the high uncertainties inherent to this task. Additionally, we observe that DUSt3R often predicts precise camera rotations but struggles with camera centers in many cases (\eg, the keyboard example in Fig.~\ref{fig:supp_vis_compare}). This observation aligns with our quantitative results for camera center evaluation, presented in Tab.~\ref{tab:co3d_rot_tra}.

\section{Sparse-to-Dense Training Details and Evaluation}
\label{sec:supp_sparse_to_dense}

\begin{table*}[!t]
    \small{
    \centering
    \begin{tabular}{ll|ccccccc|ccccccc}
\toprule
& & \multicolumn{7}{c}{Rotation Accuracy ($\uparrow$, @ 15$^\circ$)}
  & \multicolumn{7}{c}{Center Accuracy ($\uparrow$, @ 0.1)}\\
& \# of Images & 2 & 3 & 4 & 5 & 6 & 7 & 8
  & 2 & 3 & 4 & 5 & 6 & 7 & 8 \\
\midrule
\parbox[t]{2mm}{\multirow{3}{*}{\rotatebox[origin=c]{90}{Seen}}} 
& Sparse Model
  & 92.5 & 93.1 & 93.4 & 93.6 & 93.6 & 93.8 & 93.9
  & 100 & 95.4 & 92.6 & 90.9 & 89.6 & 88.8 & 88.2 \\
& Dense Model (1)
  & 90.3 & 90.7 & 90.9 & 90.8 & 90.9 & 91.0 & 90.9
  & 100 & 94.9 & 91.1 & 89.0 & 87.1 & 85.7 & 84.2 \\
& Dense Model (2)
  & 93.4 & 94.0 & 94.5 & 94.8 & 95.0 & 95.2 & 95.1
  & 100 & 95.9 & 93.6 & 92.2 & 91.2 & 90.7 & 90.2 \\
\midrule
\parbox[t]{2mm}{\multirow{3}{*}{\rotatebox[origin=c]{90}{Unseen}}}
& Sparse Model
  & 87.0 & 89.2 & 90.2 & 90.7 & 91.2 & 91.7 & 92.1
  & 100 & 90.9 & 86.3 & 83.1 & 81.0 & 79.7 & 79.2 \\
& Dense Model (1)
  & 85.9 & 86.8 & 87.4 & 87.8 & 88.5 & 88.6 & 88.8 
  & 100 & 89.1 & 83.7 & 79.7 & 77.7 & 75.5 & 74.5 \\
& Dense Model (2)
  & 90.4 & 91.2 & 92.7 & 93.0 & 93.1 & 93.3 & 93.5
  & 100 & 91.1 & 87.7 & 85.3 & 83.7 & 82.7 & 82.0 \\
\bottomrule
\end{tabular}
}
\caption{
    \textbf{Camera Rotation and Center Accuracy on CO3D at Different Training stages.} On the left, we report the proportion of relative camera rotations within $15^\circ$ of the ground truth. On the right, we report the proportion of camera centers within $10\%$ of the scene scale, relative to the ground truth. To align the predicted camera centers to ground truth, we apply an optimal similarity transform ($s$, $\mathbf{R}$, $\mathbf{t}$). Hence the alignment is perfect at $N=2$ but worsens with more images.
    }
    \vspace{-5mm}
    \label{tab:supp_sparse_to_dense}
\end{table*}

\begin{figure}[!t]
\centering
\includegraphics[width=1.0\linewidth]{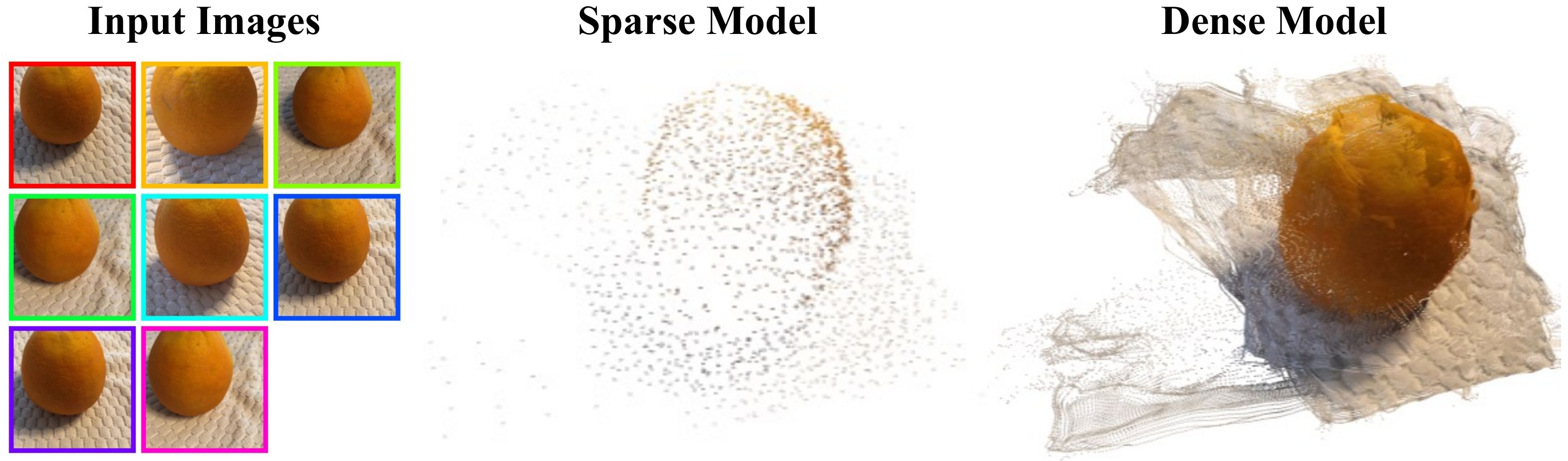}
\caption{\textbf{Qualitative Comparison of Sparse and Dense Model Outputs.} The sparse model predicts the ray origin and endpoint for each image patch, limiting its ability to capture the fine-grained details of the scene.}
\label{fig:supp_vis_sparse_vs_dense}
\end{figure}

As outlined in Sec.~\ref{sec:training_details}, we follow a sparse-to-dense strategy to train our model as we find that training the high-resolution model (\ie, the dense model) from scratch yields suboptimal performance. We visualize the output of the sparse model and dense model in Fig.~\ref{fig:supp_vis_sparse_vs_dense}. In the following, we introduce the details of training DiffusionSfM.

\noindent \textbf{Details.} Our model leverages DINOv2-ViTs14 \cite{oquab2023dinov2} as the feature backbone and takes $224\times224$ images as input. This results in $16\times16$ image patches, each with a patch size of 14. We first train a sparse model that outputs patch-wise (\ie, $16\times16$) ray origins and endpoints. Since the spatial resolution of the GT ray origins and endpoints for the sparse model aligns with the DINOv2 feature map, we use a single linear layer to embed the noisy ray origins and endpoints (without spatial downsampling), rather than a convolutional layer as shown in Eq.~\ref{eq:line_segment_embedding}. We also remove the DPT \cite{ranftl2021vision} decoder in our sparse model. Subsequently, we initialize our dense model from the pre-trained sparse model to predict dense (\ie, $256\times256$) ray origins and endpoints. We copy-paste the DiT \cite{peebles2023scalable} weights from the sparse model. Whereas for the convolutional layer used to embed ray origins and endpoints, we duplicate the linear-layer weights by $16\times16$ (as the patch size of the conv-layer is 16) and then divide them by 256 to account for the patch-wise addition. While the DiT in the dense model has learned meaningful representations, the DPT decoder is initialized from scratch. To avoid breaking the learned DiT weights in early training iterations, we freeze its weights while only training the convolutional embedding layer and the DPT decoder for a few iterations. This warm-up model is referred to as Dense Model (1). After that, we train the whole model together, including the DINOv2 encoder as well (which was frozen in the previous stage). During this phase, we apply a lower learning rate (0.1$\times$) to both the DINOv2 encoder and DiT compared to the DPT decoder. The fully trained model is referred to as Dense Model (2). We compare the performance of DiffusionSfM-CO3D at each stage in Tab.~\ref{tab:supp_sparse_to_dense}.

\begin{figure*}[!t]
\centering
\includegraphics[width=\linewidth]{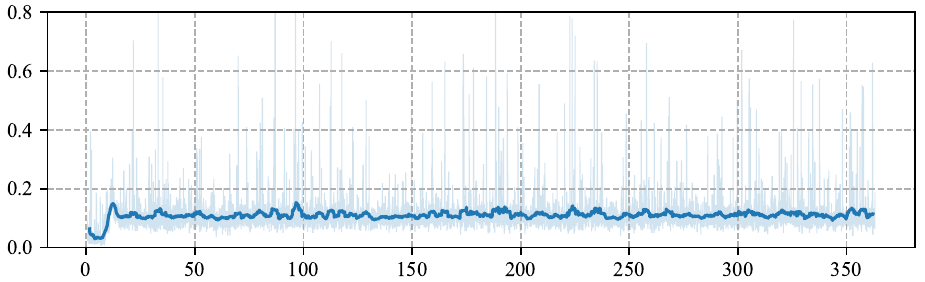}
\caption{\textbf{Training Loss Curve for DiffusionSfM without Homogeneous Representation.} The X-axis represents training iterations (in thousands, k), and the Y-axis denotes the loss value. Without incorporating a homogeneous representation for ray origins and endpoints, the model struggles to train effectively due to significant scale differences across various scene components.}
\label{fig:supp_no_homogeneous}
\end{figure*}

\noindent\textbf{Training Resources.} (1) \ours-CO3D: We train the sparse model using 4 H100 GPUs with a total batch size of 64 for 400,000 iterations, which takes approximately 2 days. To warm up the dense model, we freeze the DiT weights and train for 50,000 iterations. We then unfreeze the full model and continue training for another 250,000 iterations on 4 H100 GPUs with a batch size of 48, requiring an additional 2 days. (2) \ours: This variant is trained with 8 H100 GPUs and a larger batch size. The sparse model is trained for 1,600,000 iterations with a total batch size of 288 (the first 1,080,000 iterations are run with 4 GPUs and a batch size of 64), which takes approximately two weeks. The dense model is trained for 800,000 iterations, including 50,000 warm-up iterations, using a batch size of 96 and taking around 7 days.

\section{The Effect of Homogeneous Representation}
\label{sec:supp_homogeneous_representation}
To underscore the importance of the proposed homogeneous representation for ray origins and endpoints, we train a variant of DiffusionSfM using these components directly in $\mathbb{R}^3$ (\ie, without using homogeneous coordinates). For this model, we employ a scale-invariant loss function, as used in DUSt3R \cite{wang2023DUSt3R}. The training loss curve for this model is shown in Fig.~\ref{fig:supp_no_homogeneous}. Notably, the model fails to converge, with the training loss remaining persistently high. This failure occurs because our diffusion-based approach assumes input data within a reasonable range, as the Gaussian noise added during training has a fixed standard deviation of 1 (before scaling). Consequently, training scenes with substantial scale differences across components disrupt the model's learning process. In contrast, employing homogeneous coordinates enables the normalization of the input data to a unit norm, which not only stabilizes training and facilitates convergence but also provides an elegant representation of unbounded scene geometry.

\section{Converting Ray Origins and Endpoints to Camera Poses}
\label{sec:supp_rays_to_cameras}

The camera centers for each input image are recovered by averaging the corresponding predicted ray origins. To determine camera rotations and intrinsics, we follow the method proposed by Zhang \etal~\cite{zhang2024cameras}, which involves solving for the optimal homography that aligns the predicted ray directions with those of an identity camera. For additional details, we refer readers to Zhang \etal~\cite{zhang2024cameras}.